%% file: main.tex
\documentclass[sigconf]{acmart}
\setcopyright{none}
\renewcommand\footnotetextcopyrightpermission[1]{}
\usepackage[utf8]{inputenc} 
\usepackage[T1]{fontenc}    
\usepackage{hyperref}       
\usepackage{url}            
\usepackage{booktabs}       
\usepackage{amsfonts}       
\usepackage{nicefrac}       
\usepackage{microtype}      
\usepackage{xcolor}         
\usepackage{makecell}
\usepackage{multirow}
\usepackage{float}
\usepackage{threeparttable}
\usepackage{enumitem}
\usepackage{ulem}

\usepackage{wrapfig}
\usepackage{newfloat}
\usepackage{listings}
\usepackage{bbding}
\usepackage{fancyhdr}
\usepackage{xcolor}
\usepackage{tablefootnote}
\usepackage{algorithm}
\usepackage{algpseudocode}
\usepackage{stfloats}
\usepackage{bm}
\usepackage{graphicx}
\usepackage{comment}
\usepackage{amsmath}
\AtBeginDocument{%
  }
\setcopyright{acmlicensed}
\copyrightyear{2018}
\acmYear{2018}
\acmDOI{XXXXXXX.XXXXXXX}

\acmConference[Big Data Mining and Analytics '25]{Make sure to enter the correct
  conference title from your rights confirmation emai}
\acmISBN{978-1-4503-XXXX-X/18/06}

\begin{document}
\newcolumntype{L}[1]{>{\raggedright\arraybackslash}p{#1}}
\newcolumntype{C}[1]{>{\centering\arraybackslash}p{#1}}
\newcolumntype{R}[1]{>{\raggedleft\arraybackslash}p{#1}}
\title[Exploring Knowledge Integration and Controllable Generation on Large Language Models]{Prompting is not Enough: Exploring Knowledge Integration and Controllable Generation}

\author{Tingjia Shen}
\affiliation{%
  \institution{University of Science and Technology of China}
  \city{Hefei}
  \state{Anhui}
  \country{China}}
\email{jts_stj@mail.ustc.edu.cn}

\author{Hao Wang}
\authornote{Corresponding author.}
\affiliation{%
  \institution{University of Science and Technology of China}
  \city{Hefei}
  \state{Anhui}
  \country{China}}
\email{wanghao3@ustc.edu.cn}

\author{Chuan Qin}
\authornotemark[1]
\affiliation{%
  \institution{Computer Network Information Center, Chinese Academy of Sciences}
  \city{Beijing}
  \country{China}}
\affiliation{%
  \institution{Chinese Academy of Sciences}
  \city{Beijing}
  \country{China}}
\email{chuanqin0426@gmail.com}

\author{Ruijun Sun}
\affiliation{%
  \institution{University of Science and Technology of China}
  \city{Hefei}
  \state{Anhui}
  \country{China}}
\email{rjsun@mail.ustc.edu.cn}

\author{Yang Song}
\affiliation{%
  \institution{BOSS Zhipin Career Science Lab}
  \city{Beijing}
  \country{China}}
\email{songyang@kanzhun.com}

\author{Defu Lian}
\affiliation{%
  \institution{University of Science and Technology of China}
  \city{Hefei}
  \state{Anhui}
  \country{China}}
\email{liandefu@ustc.edu.cn}

\author{Hengshu Zhu}
\affiliation{%
  \institution{Computer Network Information Center, Chinese Academy of Sciences}
  \city{Beijing}
  \country{China}}
\affiliation{%
  \institution{Chinese Academy of Sciences}
  \city{Beijing}
  \country{China}}
\email{zhuhengshu@gmail.com}

\author{Enhong Chen}
\affiliation{%
  \institution{University of Science and Technology of China}
  \city{Hefei}
  \state{Anhui}
  \country{China}}
\email{cheneh@ustc.edu.cn}

\renewcommand{\shortauthors}{}

\renewcommand{\shortauthors}{}

\begin{abstract}
Open-domain question answering (OpenQA) represents a cornerstone in natural language processing (NLP), primarily focused on extracting answers from unstructured textual data. With the rapid advancements in Large Language Models (LLMs), LLM-based OpenQA methods have reaped the benefits of emergent understanding and answering capabilities enabled by massive parameters compared to traditional methods. However, most of these methods encounter two critical challenges: how to integrate knowledge into LLMs effectively and how to adaptively generate results with specific answer formats for various task situations. To address these challenges, we propose a novel framework named \textbf{GenKI}, which aims to improve the OpenQA performance by exploring \underline{\textbf{K}}nowledge \underline{\textbf{I}}ntegration and controllable \underline{\textbf{Gen}}eration on LLMs simultaneously. Specifically, we first train a dense passage retrieval model to retrieve associated knowledge from a given knowledge base. Subsequently, we introduce a novel knowledge integration model that incorporates the retrieval knowledge into instructions during fine-tuning to intensify the model. 
Furthermore, to enable controllable generation in LLMs, we leverage a certain fine-tuned LLM and an ensemble based on text consistency incorporating all coherence, fluency, and answer format assurance. 
Finally, extensive experiments conducted on the TriviaQA, MSMARCO, and CMRC2018 datasets, featuring diverse answer formats, have demonstrated the effectiveness of GenKI with comparison of state-of-the-art baselines.
Moreover, ablation studies have disclosed a linear relationship between the frequency of retrieved knowledge and the model's ability to recall knowledge accurately against the ground truth. Tests focusing on the out-of-domain scenario and knowledge base independence scenario have further affirmed the robustness and controllable capability of GenKI. Our code of GenKI is available at~\href{https://github.com/USTC-StarTeam/GenKI}{\textcolor{blue}{https://github.com/USTC-StarTeam/GenKI}}.
\end{abstract}

\keywords{Open-Domain Quesion Answering, Large Language Model}
\maketitle
\input{chapter1_intro.tex}

\input{chapter2_RelatedWork}

\input{chapter3_Definition}

\input{chapter4_Methodology}
\input{chapter5_ExperimentalSetup}

\input{chapter6_Results}

\input{chapter7_Conclusion}

\vskip 2mm
\large
\noindent
\textbf{Acknowledgment}
\vskip 2mm

\noindent
This research was funded by the National Natural Science Foundation of China (U23A20319, 62441227, 62441239, 62472394, 62202443), the Anhui Province Science and Technology Innovation Project (202423k09020011), and the Anhui Provincial Science and Technology Major Project (2023z020006).
\bibliographystyle{plain}
\bibliography{BigData}

\appendix

\end{document}

%% file: chapter1_intro.tex

\section{Introduction}

Open-domain question answering (OpenQA) has garnered significant attention in the Natural Language Processing (NLP) community, as it offers enhanced user-friendliness and efficiency compared to conventional search engines.
OpenQA seeks to respond to questions utilizing auxiliary knowledge sources~\cite{zhu2021retrieving}, necessitating models that possess both knowledge capability and comprehension abilities.
Conventional methods usually entail a retriever-reader framework~\cite{DBLP:conf/iclr/DasDZM19}. The evolution of retriever models has progressed from BM25~\cite{10.1561/1500000019} and TF-IDF~\cite{aizawa2003information} to dense-vector-based approaches, such as DPR~\cite{karpukhin-etal-2020-dense} and SEAL~\cite{DBLP:conf/nips/BevilacquaOLY0P22}. Concurrently, reader models have diversified, spanning from extractive readers like DPR-reader~\cite{cui-emnlp2019-cmrc2018} to generative readers, such as FiD~\cite{DBLP:journals/corr/abs-2007-01282} and RAG~\cite{DBLP:conf/nips/LewisPPPKGKLYR020}, targeting span and free answering tasks respectively. 
Recently, the rapid development of Large Language Models (LLMs) has motivated researchers to incorporate them into OpenQA, driven by the models' advanced abilities in recommendation~\cite{wang2021hypersorec} and natural language reasoning~\cite{shen2024exploring}.
For instance, ICL~\cite{li2023selfprompting} uses certain prompts towards better integration of external knowledge with in-context learning, while Replug~\cite{shi2023replug} enhances an LLM with retrieved knowledge as prompts.
 
\begin{figure}[t]
  \includegraphics[width=0.98\linewidth]{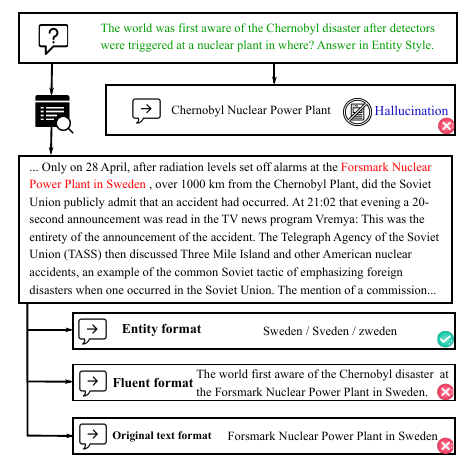}
  \caption{An illustration of hallucination encountered when employing LLMs in OpenQA task, along with the variations in answer formats across different OpenQA datasets.}
  \label{fig:GenKI_intro}
\end{figure}

However, when utilizing language models for OpenQA, two notable error phenomena persist, as depicted in Fig.~\ref{fig:GenKI_intro}. These challenges include:
Challenge (1): \textbf{knowledge deficiency}. When prompted to identify ``which nuclear plant's detectors first alerted the world to the Chornobyl disaster'', the LLM fabricates an incorrect response. This issue stems from the limited memorization capacity of LLMs. Specifically, LLMs not only face challenges in retaining less frequent knowledge~\cite{kandpal2023large}, but also tend to produce hallucinations~\cite{DBLP:conf/emnlp/0001PCKW21} when the requisite knowledge is absent from their pretraining data. Existing methods primarily address this issue through pre-training~\cite{DBLP:journals/corr/abs-2204-02311} or prompting~\cite{ren2023investigating}. Pre-training is too broad and lacks precision while prompting overlooks the model's potent knowledge storage capabilities~\cite{petroni-etal-2019-language}. Challenge (2): \textbf{answer format alignment}. Various OpenQA datasets demonstrate unique answer formats, exemplified by the three markedly different formats depicted in Fig.~\ref{fig:GenKI_intro}. 
However, even when tasked to produce answers in a specific format, the outputs of LLMs often diverge from these expected formats. This discrepancy is largely due to the distributional biases present in the pre-training corpora, as discussed in~\cite{kandpal2023large}.

Presently, to tackle challenge (1), PaLM 540B~\cite{DBLP:journals/corr/abs-2204-02311} enlarges the model scale to accommodate a greater wealth of knowledge, which, however, leads to unacceptable computational complexity.
So the ideology of RAG~\cite{gao2023retrieval} integrating knowledge with prompts for efficiency has raised. After researchers recognize the significance of retrieving knowledge, they seek to enhance the quality of retrieved content. Typical examples are enabling large models to autonomously assess the quality of retrieved content for answering questions~\cite{ren2023investigating}, or iteratively conducting multi-step retrievals to progressively enhance the quality of retrieved content and align with the preferences of LLM\cite{asai2023self}. Nevertheless, integrating knowledge through prompt-based approaches treats LLMs merely as instruction interpreters, disregarding their potent knowledge storage capabilities~\cite{petroni-etal-2019-language}. Consequently, this approach yields unstable and suboptimal results. Toward challenge (2), InstructGPT~\cite{ouyang2022training} and FLAN~\cite{DBLP:conf/iclr/WeiBZGYLDDL22} adopt a pre-training and instruction tuning paradigm and attempt to encapsulate all instructions along with the language understanding capabilities equipped with knowledge at the same time. Nevertheless, this paradigm requires the storage of massive real-world data~\cite{yin2024dataset}, and the computational burden is quadratic with respect to the number of model parameters~\cite{hoffmann2022training}. To address this issue, CoF-CoT~\cite{nguyen2023cof} attempts to utilize a chain of instruction to control the output format. However, a distribution gap persists between the formatted output and knowledge. Consequently, the collaborative training procedure is prone to encountering contradictions~\cite{DBLP:conf/acl/MallenAZDKH23}.

In general, although the aforementioned generator enhancement methods have achieved remarkable progress in the OpenQA field, current pre-training or fine-tuning-based methods only focus on either full knowledge adaption or applying LLM on both reasoning and answer alignment missions, considering two different kinds of task in one stage of tuning. This leads to distribution misalignment and limitations in sufficient knowledge integration and controllable generation performance in different task situations.



 Towards these challenges, in this paper, we propose a novel framework to improve the OpenQA performance by exploring knowledge integration and controllable generation on LLMs, namely GenKI. 
 We introduced a novel three-stage paradigm for retrieval, knowledge integration, and controllable generation instead of the two-stage paradigm in Retrieval and Generation (RAG), ensuring that the model focuses on either knowledge integration or controlled generation in each instance, to avoid the distributional differences brought by these two tasks.
 Specifically, in the knowledge integration model, rather than adopting normal fine-tuning-based method, we propose an innovative method of LLM combining autoregressive training loss and supervised fine-tuning loss to involve the retrieval results. In comparison to the prior research on in-context learning using RAG or fine-tuning methods, which solely instruct the LLM, our model acquires new knowledge more stably by storing the retrieved domain knowledge within parameters rather than relying on prompts.
 In the controllable generation model, we first utilize a certain fine-tuned LLM to post-process the generated answer towards a certain format. Subsequently, we propose a novel ensemble methodology based on text consistency to incorporate both coherence and fluency from the reward model and faculty from the external selection, thereby ensuring adaptive alignment between the output of our model and the target format. 

Extensive experiments on three benchmark datasets, including TriviaQA, MSMARCO, and CMRC2018, have demonstrated the effectiveness of the proposed framework.
Additionally, our ablation study results reveal: (1) a linear relationship between the quality of retrieved results and the model's knowledge proficiency, and (2) the impact of different structures on our answers, with the tuning mechanism playing a key role in generating answers in a specific format, and the reward model excelling in choosing fluent sentences.
The main contributions are summarized as follows:
\begin{itemize}
    \item We propose a novel problem on the phenomena of knowledge deficiency and answer format misalignment of LLMs on the OpenQA task. 
    \item We introduce a novel three-stage paradigm of retrieval, knowledge integration, and controllable generation, avoiding the distributional differences of LLM in two-stage RAG, and propose a novel ensemble method based on text consistency, ensuring both coherence and faculty of the generated answer.
    \item Extensive experiments on three benchmark datasets and analysis on OOD situation and independence knowledge base tests unequivocally affirm the distinguished controllable generation and robustness of our framework. 
    \item Through analysis on ablation experimental results, we find the linear fitting relationship between the quality of retrieved results and the knowledge proficiency of the model and a way to judge whether the model is pre-trained on knowledge demanded, for future research on knowledge integration in LLMs.
\end{itemize}

%% file: chapter2_RelatedWork.tex
\section{Related Work}

\subsection{Open-domain Question Answering}

OpenQA is a crucial task in NLP, which aims to answer information-seeking questions based on auxiliary knowledge source~\cite{gu2025rapid}. Traditionally, an OpenQA approach entails a retriever-reader framework, with the retriever used to locate relevant knowledge and the reader generating the response. Extensive research has delved into retriever techniques, encompassing a range from traditional ones such as BM25~\cite{10.1561/1500000019} and TF-IDF~\cite{aizawa2003information} to dense-vector-based methods like DPR~\cite{karpukhin-etal-2020-dense}, as well as retrievers like SEAL~\cite{DBLP:conf/nips/BevilacquaOLY0P22}, which leverages all n-grams in a passage as its identifiers. Since then Iterative retrievers such as GRAPHRETRIEVER~\cite{DBLP:journals/corr/abs-1911-03868} and GAR~\cite{mao-etal-2021-generation} have been raised for answering complex questions like those requiring multi-hop reasoning task~\cite{yang2018hotpotqa}. Reader models exhibit a spectrum ranging from extractive readers to generative readers. Extractive readers develop from origin DPR-reader~\cite{cui-emnlp2019-cmrc2018} to graph-based readers
like GRAPHREADER~\cite{DBLP:journals/corr/abs-1911-03868} and readers ensembling methods like BERTserini~\cite{DBLP:conf/naacl/YangXLLTXLL19} focus on span-based text answering~\cite{cui-emnlp2019-cmrc2018}. In contrast, generative readers such as FiD~\cite{DBLP:journals/corr/abs-2007-01282} and RAG~\cite{DBLP:conf/nips/LewisPPPKGKLYR020} focus on free-form text answering~\cite{bajaj2018ms,joshi-etal-2017-triviaqa}. 
Certainly, given the impressive capabilities showcased by LLMs, developed after generative readers, traditional approaches have witnessed a shift in prominence, often assuming supplementary roles alongside these models. However, valuable methods can still be gleaned from these works. In our study, one approach, namely Dense Passage Retrieval (DPR), was integrated as a retrieval plugin.


\subsection{LLMs-based Open-domain Question Answering}
Recent years have witnessed a surge in interest in LLMs for their scaling~\cite{shen2025optimizingsequentialrecommendationmodels} capabilities across various NLP tasks like Recommendation~\cite{wu2024survey} and NER~\cite{min2021recent}. Amidst this trend, handling OpenQA based on LLMs is emerging as a popular research direction. Research in this domain can be mainly divided into two categories: discriminative language models-based approaches and generative language models-based approaches. In the first type of approaches, researchers typically fine-tune BERT~\cite{devlin2018bert} or RoBERTa~\cite{liu2019roberta} to build a reader aligned to answering tasks~\cite{joshi2020spanbert,xiao-etal-2021-ernie,yasunaga-etal-2022-linkbert}.
Following the occurrence of generative language models, such as GPT~\cite{brown2020language}, GLM~\cite{du2021glm}, LLaMA~\cite{touvron2023llama}, the researchers began adopting these models for OpenQA. This shift was driven by their proven capability to handle OpenQA without relying on retrievers~\cite{wei2021finetuned}. Nevertheless, Self-Prompting~\cite{li2023selfprompting} persists in trying to enhance LLM itself by in-context learning. There is also some research exploring the role of retrievers in enhancing the performance of generative large models on these tasks, such as REPLUG~\cite{shi2023replug} and kNN-LM~\cite{DBLP:conf/emnlp/ZhongLC22}. Afterward, researchers recognized the effectiveness of the paradigm using retrieved knowledge, i.e., RAG~\cite{gao2023retrieval}, and began to work on further refining this paradigm.

\subsection{RAG-LLMs on Open-domain Question Answering}
Advanced RAG~\cite{gao2023retrieval,jiang2024enhancing} has made improvements to overcome the deficiencies of Naive RAG. The current emphasis of the advanced RAG model is on enhancing the retriever and generator components. 
On the retriever enhancement, LangChain~\cite{topsakal2023creating}
first proposes a modularized retrieval and reranking framework. RRR~\cite{ma2023query} and AAR~\cite{yu2023augmentation} try to align Retriever with LLM using modularized or end-to-end structure. Finally, the LLaMA-index{\hspace{-10mm}\footnote{\noindent https://www.llamaindex.ai}} proposed an end-to-end alignment and rerank structure to merge them.
On the Generator enhancement, end-to-end frameworks~\cite{sachan2021end} and Pre-training methods like PaLM~\cite{chowdhery2022palm} are proposed for pre-training LLMs adapting to downstream knowledge. Nevertheless, with the expanding scale of LLMs, the viability of the pre-training method diminishes over time. So recently prompt-engineering~\cite{li2023selfprompting} and supervised fine-tuning\cite{ouyang2022training} are the technologies widely used. RFiD~\cite{wang2023rfid} finetunes the decoder to generate answers by relationships and features. GAG~\cite{abdallah2023generator} finetunes the model for both answering and generating contextually rich documents tailored to the given question. Self-rag~\cite{asai2023self}, iGFT~\cite{tong2025missteps} and SeaKR~\cite{yao2024seakrselfawareknowledgeretrieval} further finetune an LLM to generate and reflect on retrieved passages and its own generations using special retrieval or rerank tokens. 
However, these generator-based methods merely input retrieved knowledge, which is more precise than pre-training, as prompts into the model. As mentioned in our introduction, the knowledge provided by prompts is not stable.
Overall, although the aforementioned generator enhancement methods have achieved remarkable progress in the OpenQA field, current pre-training or fine-tuning-based methods only focus on either full knowledge adaption or applying LLM on both reasoning and answer alignment missions. This leads to limitations in sufficient knowledge integration and controllable generation performance in different task situations, which is the area we aim to address.

%% file: chapter3_Definition.tex
\section{Preliminary}
\subsection{OpenQA Problem Definition}



Given a domain knowledge base (e.g., Wikipedia) comprising a set of passagess $P=\{p_{1},p_{2},\ldots,p_{n_1}\}$, a specified answer format request $F$, and each query question $q$ within query set $Q=\{q_1,q_2,\ldots,q_{n_2}\}$ with $n_1$ and $n_2$ representing numbers of passages $P$ and query set $Q$. OpenQA aims to train a model $M$ that can extract relevant knowledge and generate the ideal answer $a_i = M(q_{i},F,P)$ in response to each query $q$ with index $i\in[1,2,\ldots,n_2]$ with format $F$, guided by the information in the knowledge base.

In authentic knowledge bases like Wikipedia, information is typically structured within web pages or documents. By segmenting it into sentence form, we can derive the set of sentences denoted as~$P$. Take, for instance, the question $q_{i}$ is ``Where was the initial awareness of the Chernobyl incident triggered?''. Given that the desired format $F$ is entity style, the model is expected to leverage insights extracted from these sentences to generate an answer such that $a_i=M(q_{i}, F, P)=$``Sweden''.

\subsection{Instruct tuning on LLMs}
In the OpenQA pipeline with LLMs, we employ a fine-tuned LLM denoted as the model $M$ for answer formulation. The fine-tuning process of the LLM facilitates its adaptation to the distribution and domain knowledge relevant to downstream tasks. This process essentially encompasses an equivalent final objective loss, resembling that of autoregressive training, the original training loss $\mathcal{L}_f$ is outlined as follows:
\begin{equation}
\mathcal{L}_f=\mathop{\mathrm{max}}_{\Phi}\sum_{x_i, a_i\in T}\sum_{t=1}^{|a_i|}\mathrm{log}(\ \mathop{\mathrm{Prob\ }}_{\Phi}(a_{i}[t]|x,a_{i}[<t])),
\end{equation}
where $\Phi$ represents the parameters of LLM to be optimized, $T$ denotes the training set, $x$ refers to the input context encompassing both an instruction and a query question,  $Prob(\cdot)$ denotes the generation probability, $a_{i}[t]$ is the $t$-th token ,$||$ denotes the total number of a set while $a_{i}[<t]$ denotes token sequence before the $t$-th token of the generated answer word.
For all tunings of our experiment, we adopt the LoRA~\cite{hu2021lora} method of lightweight fine-tuning to reduce the required GPU memory consumption.

%% file: chapter4_Methodology.tex
\section{Methodology}

\begin{figure*}[t]
  \centering
  \includegraphics[width=0.95\linewidth]{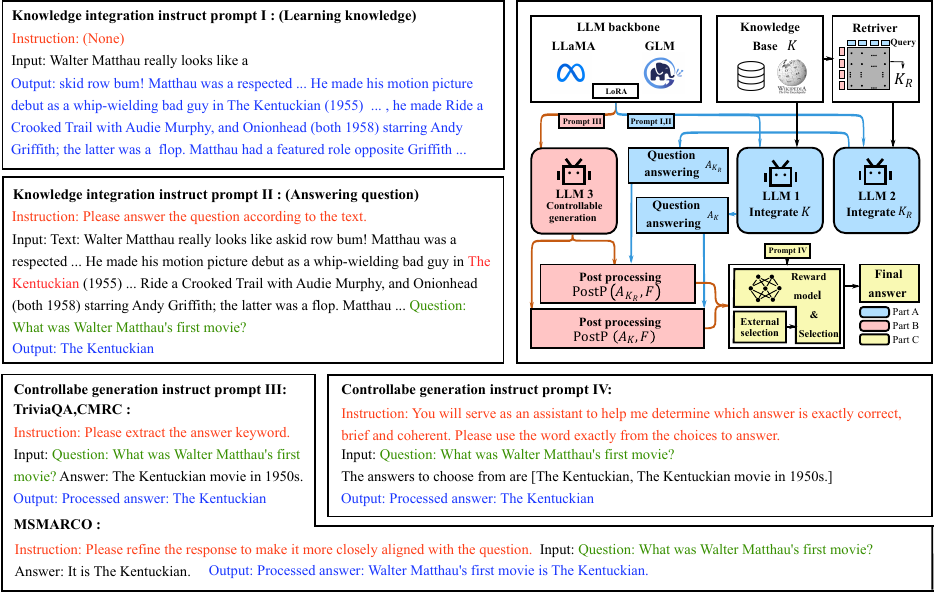}
  \caption{Overall framework of GenKI. The left part presents several examples of our prompts used in the knowledge integration and controllable generation module. The right part outlines the process of tuning LLMs to learn knowledge and answer questions in specific task situations.}
  \label{fig:GenKI_structure}
\end{figure*}

\noindent To address the knowledge deficiency and uncontrollable generation issues in LLMs for OpenQA, our system combines three modules as illustrated in Fig.~\ref{fig:GenKI_structure}. 
On the whole, we decompose the model $A=M(q_i,F,P)$ into two stage to ensure that the model focuses on either knowledge integration or controlled generation in each instance, avoiding the distributional differences brought by these two tasks. 
Specifically, we first train
a knowledge retriever module $R(P, q_i) = P_R$ to retrieve knowledge according to the query supplying the model during knowledge integration steps. Subsequently, once the retrieved knowledge is synthesized, 
a knowledge integration module $K(q_i, P_R) = A_K$ (Part $\textsl{A}$ in Fig.~\ref{fig:GenKI_structure}) is deployed to generate answers by integrating the retrieved knowledge. 
Finally, a controllable generation module involves using a fine-tuned LLM on the target dataset format for post-processing the results, which is denoted as $\mathrm{PostP}$ and a Reward model $\mathrm{Rm}$ to further ensure that the output of our method is controlled. The function of each module will be introduced in detail below. 

\subsection{Knowledge retriever}

Firstly, given vast knowledge repositories, we should prioritize enabling our model to learn from them. However, feeding all the sentences $P$ from the knowledge base into the LLM may potentially result in the model's inability to grasp essential focal points.
Therefore, we need a retriever $R(P, q) = P_R \subset P$ to retrieve knowledge $P_R$ as the initial input provided for the next knowledge integration module.

Specifically, we use a Dense Passage Retriever (DPR) to retrieve knowledge from the sentence knowledge base $P$ according to the question.
Setting the encoder as $f$, and the trainable parameters as $\Phi$, DPR computes question-sentence similarity $S$ for a given question $q$ and a set of sentences $P$ using the formula below:
\begin{equation}
S_{q,\ p,\ \Phi}=f_q(q,\Phi_q)^{\mathrm{T}}f_p(p;\Phi_p), \ p\in P 
\end{equation}
where $\Phi=[\Phi_q,\Phi_p]$ denotes the retriever question and passage encoder parameters. We choose the final Top-k passage $P_R$ as below:
\begin{equation}
  \begin{aligned}
P_R=R(P, q)=\mathrm{Top}_k(\mathrm{min}\ (S_{q,\ p,\ \Phi}, p\in P))
  \end{aligned}
\end{equation}

The retrieved result in $P_R$ will be used as the input and tuning material for knowledge integration, which can provide LLMs with more accurate knowledge.

\subsection{Knowledge integration}
\noindent From the previous part, we acquire high-quality knowledge utilizing the retriever. To integrate this high-quality knowledge into the parameters of our model, tuning is an effective way. However, the original instruction tuning is inadequate for the OpenQA task, which is attributed to its limited scope of only aiding LLMs in comprehending the objectives of different tasks. 
Hence the knowledge integration $A_K=K(q_i,P)$ aims to enable the model to receive the necessary knowledge for generating the target answer. The detailed demonstration is clearly illustrated in Fig.~\ref{fig:GenKI_structure}.
Inspired by how pretraining enables LLMs to grasp knowledge~\cite{brown2020language,du2021glm,touvron2023llama}, we introduce the following tuning optimization to enable the model to learn more domain knowledge precisely. 
However, as one of the innovations of this paper, our knowledge integration module is neither learning all knowledge extensively like pre-training nor providing knowledge only in the prompt like contextual learning. The fine-tuning paradigm designed in this paper integrates retrieved refined knowledge into LLM through fine-tuning, ensuring the accuracy and domain specificity of the knowledge provided for LLM.

Specifically, the tuning loss $\mathcal{L}_r$ of an LLM on domain passage $P$ is defined as follows:
\begin{equation}
\mathcal{L}_r=\mathop{\mathrm{max}}_{\Phi}\sum_{p_i\in P}\sum_{t=1}^{|p_i|}\mathrm{log\ }(\mathop{\mathrm{Prob\ }}_{\Phi}(p_i[t]|p_i[<t]))
\end{equation}
where || denotes the nomber of tokens of $p_i$.

The final optimization loss is as follows:
\begin{equation}
\begin{aligned}
&\mathcal{L}=\lambda_1\mathcal{L}_{r}+\lambda_2\mathcal{L}_{f}=\\
&\mathop{\mathrm{max\ }}_{\Phi_L}(\lambda_1\sum_{p\in P}\sum_{t=1}^{|p_i|}\mathrm{log\ }(\mathop\mathrm{{Prob\ }}_{\Phi+\Phi_L}(p_i[t]|p_i[<t])))+ \lambda_2\mathcal{L}_{f} ,
\end{aligned}\label{PSqrt} 
\end{equation}

\begin{figure}[t]
  \includegraphics[width=\linewidth]{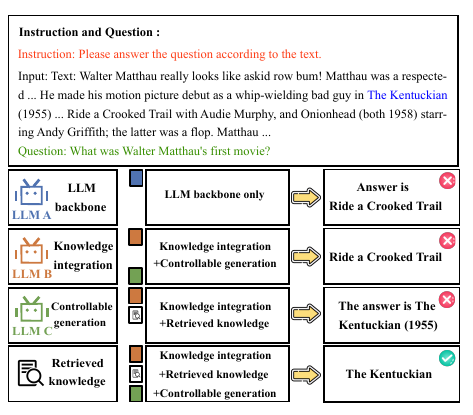}
  \caption{Real cases of outputs after Knowledge integration and controllable generation structure.}
  \label{fig:GenKI_output}
\end{figure}
\noindent where $\mathcal{L}_{f}$ is the loss of original instruction tuning, $\Phi_L$ is the LoRA parameters and we only update LoRA parameters during the training process. The prompts we used are shown in Fig.~\ref{fig:GenKI_structure}. In the experiment, we set $\lambda_1>\lambda_2>0$ to make the model focus on knowledge without losing the ability to answer questions.


Given that retrieval can occasionally result in inaccuracies, leading to errors in $A'=K(q,P_R)$, where $A'$ is the answer adopting retrieved passages $P_R=R(P,q_i)$, it is necessary to retain the option of using the full passage $P$ without retrieval. Therefore, we utilize this module with inputs either from the complete passage $P$ or from the retrieved knowledge $P_R$, using $A=K(q,P)$ or $A'=K(q,P_R)$, respectively. These output options are then presented to the subsequent controllable generation step, allowing for a selective choice.

Incorporating knowledge during finetuning using this method leverages the advantages of the retriever, enabling the model to learn more precise and specialized knowledge. By storing domain knowledge in LLMs' parameters, our work leads the model to learn knowledge more stably, resulting in superior performance. 


\subsection{Controllable generation}
\noindent From the knowledge integration model, we gain a draft of the answer. While the draft of this answer already contains all the knowledge needed to address the question, it still does not meet our goal of formatted generation. 
In order to align the model's output with the dataset's answers and our expectations, we need to introduce another framework for post-processing the answers from the last module.

The approach employed for achieving controllable generation utilizes the identical loss function as instruction tuning. 
However, due to the fact that the preceding step has yielded answers imbued with high-quality knowledge, aligning them into the format we require becomes a more manageable task than instruction tuning for the model to undertake. After fine-tuning the model with a certain amount of training data, we obtain the post-processing model\ $\mathrm{PostP}$. The instruction we used for fine-tuning is demonstrated in Part {B} in Fig.~\ref{fig:GenKI_structure}.

Additionally, derived from the previous step, we have acquired two distinct answers, denoted as $\mathrm{PostP}(A_K, F)$ and $\mathrm{PostP}(A_{K_R}, F)$. 
The imminent task involves the meticulous selection of one among these to serve as the ultimate output. 
In order to make our answer reliable and controllable, we combine a reward model to effectively select answers that match the output style of the dataset and an external judging model when these two answers are severely different. The pair-wise loss $\mathcal{L}_{R_m}$ of the reward model chooses the best answer between two answers, formulated as below:
\begin{equation}
\mathcal{L}_{Rm}=-\mathrm{log\ }(\sigma(Rm(A_+,F)-Rm(A_-,F))
\end{equation}
where $Rm(A)$ is the average score of the embedding of the final token of $A$, $A_+$ and $A_-$ denotes the positive and negative answer respectively. 
However, relying solely on the reward model ensures that the model aligns with the preference for the target format $F$. To ensure coherence and consistency in the format of the output $A$, we need to introduce an additional model for analysis and filtering.
Specifically, we then design the judgment criteria score $S_c$ as below, encouraging answer format alignment and coherence enhancement when answers are long and similar:
\begin{equation} 
  \begin{aligned}
  S_c=&\mathrm{exp\ }(|C_s(QA_1)-C_s(QA_2)|)-\\
  &\frac{|R_m(A_1,F)-R_m(A_2,F)|}{\overline{R_m(A,F)}\overline{\mathrm{len\ }(A)}})
\end{aligned},
\end{equation}
where $Q$ denotes questions, $A_1$ and $A_2$ represents $\mathrm{PostP}(A_K)$ and $\mathrm{PostP}(A_{K_R})$ respectively. $\mathrm{len\ }(A)$ is the length of answers and $\overline{\mathrm{len\ }(A)}$ 
 is the mean of $A$, In this example, \(A = \{A_1, A_2\}\). Operations like $\mathrm{exp}(\cdot)=e^{\cdot}$ aim to make consistency score $C_S$ and reward score $Rm$ at the same scale, while $\mathrm{len\ }(A)$ is designed to make the model focus more on fluency when the answer is longer. We adopt Normalized Inverse Sentence Frequency (NISF)\cite{ke-etal-2022-ctrleval} to calculate answer consistency $C_S$, which is formally calculated as below:
 \begin{equation}
  \begin{aligned}
C_s(Q, A)=&\mathrm{NISF\ }(A) \mathrm{log\ }(\mathrm{Prob}_\theta(A|Q[\mathrm{Mask}]))\\
&+\mathrm{NISF\ }(Q) \mathrm{log\ }(\mathrm{Prob}_\theta(Q|[\mathrm{Mask}]A))
  \end{aligned}
\end{equation}
\begin{equation}
\mathrm{NISF\ }(Y_j)=\frac{\mathrm{ISF}(Y_j)}{\Sigma_{k=1}^{M}\mathrm{ISF}(Y_k)},
\end{equation}
where $Q$ and $Y=\{Y_1,Y_2,\ldots, Y_j\}$ denote the question and the sentences of the question-answer pair, $[\mathrm{Mask}]$ denotes masking to calculate probability using LM and $A$ denotes the answer. $\mathrm{ISF\ }(Y)$ is the inverse sentence frequency of $Y$, which is related to the maximum of maximum inverse word frequency $(\mathrm{IWF})$ below,
\begin{equation}
\mathrm{IWF}(\omega)=\frac{\mathrm{log\ }(1+\mathrm{Count})}{f_\omega}
\end{equation}
\begin{equation}
\mathrm{ISF}(Y_j)=\mathop{\mathrm{max}}_{\omega\in Y_j}\mathrm{IWF}(\omega)
\end{equation}
where $\{\omega_{11},..., \omega_{1i},..., \omega_{ji}\}$ denotes the words of the sentences. $\mathrm{Count}$ indicates the total number of sentences, $f_\omega$ denotes the frequency of $\omega$. After computing the score, we can employ the reward model to ensure answer coherence and format alignment or prioritize the correctness and reliability of answers derived through $\mathrm{Ext}$, an external selection using chatGPT. Despite chatGPT's performance not exceeding our results from the previous stage, it plays a role in augmenting overall capability. Details are presented in Section 6.3. The overall algorithm is delineated in Algorithm ~\ref{alg:GenKI}.

\begin{algorithm}[t]
\renewcommand{\algorithmicrequire}{\textbf{Input:}}
\renewcommand{\algorithmicensure}{\textbf{Output:}}
\caption{Overall pseudo code of GenKI}
\begin{algorithmic}[1]
\Require Knowledge base $K$, Retriever $R$, LLM $L_1,\ L_2,\ L_3$, question set $Q$
\Ensure Answer $A$ 
\State Train LLM $L_1$ and $L_2$ with $K$ and $K_R=R(K,Q)$ respectively  
\Comment{Prompt I, II is used here}
\State Gain format passage $P_f$ using $L_2$ answering training queries
\State Train LLM $L_3$ with $P_f$\ \ \ \ \ \ \ \ \ \ \ \ \ \ \Comment{Prompt III is used here}
\State $A_K=L_1(Q),A_{K_R}=L_2(Q);$
\State $A_1=\mathrm{PostP}(A_K,F)=L_3(A_K,F),A_2=\mathrm{PostP}(A_{K_R},F)=L_3(A_{K_R},F);$
\State $S_c=\mathrm{exp}(|C_s(A_1)-C_s(A_2)|)-\frac{|R_m(A_1)-R_m(A_2)|}{\overline{R_m(A)}\overline{\mathrm{len\ }(A)}};$
\If{$S<0$}
    \State $A=\mathop{\mathrm{max}}_{A}(\mathrm{Rm}(\mathrm{PostP}(A_K,{F})),\mathrm{Rm}(\mathrm{PostP}(A_{K_R},{F})));$
\Else
    \State $A=\mathrm{Ext\ }(\mathrm{PostP}(A_K,{F}),\mathrm{PostP}(A_{K_R},F));$\newline
    \ \ \ \ \ \ \ \ \ \ \ \ \ \ \ \ \ \ \ \Comment{Prompt IV is used here}
\EndIf
\end{algorithmic}
\label{alg:GenKI}
\end{algorithm}

By combining instruction-tuned LLM and a controllable answer selection model, we gain an answer with both fluency brought by the reward model and answer-format accuracy brought by instruction-tuned LLM.
Finally, to provide a clearer demonstration of how different components of GenKI contribute to its performance, we present real cases of GenKI’s outputs at different stages in Fig.~\ref{fig:GenKI_output}. 
This visualization allows us to observe GenKI’s transformation and refinement of outputs as it progresses through each stage of the process.

%% file: chapter5_ExperimentalSetup.tex
\section{Experimental Settings}
\noindent We present the datasets, baselines, and evaluation metrics for comparing with baselines and evaluating the effectiveness of our model below.

\subsection{Datasets} 

\noindent To demonstrate the performance of our proposed model on different answer formats of OpenQA, we conduct experiments on three publicly aware datasets, TriviaQA~\cite{joshi-etal-2017-triviaqa}, MSMARCO~\cite{bajaj2018ms}, and CMRC-2018~\cite{cui-emnlp2019-cmrc2018}, which not only represent different answering format (free answer or span answer from the passage) but also include various answering lengths (one or two words or a sentence). 
In terms of data pre-processing, we filter out the data without correct answers on the MSMARCO dataset and randomly sample the same amount of data as TriviaQA to comparably evaluate performances. The detailed description and statistics of these datasets\footnote{\noindent The processed dataset will be open once accepted.} can be referred to in table~\ref{tab:Datasetdescribe}.

Specifically, TriviaQA is a reading comprehension dataset containing over 650, 000 training question-answer-evidence triples. The answers are mostly composed of entities in Wikipedia, containing one or two words. MSMARCO dataset contains 8, 841, 823 passages from documents retrieved by the Bing search engine, which provides the necessary information for curating the natural language answers. Different from TriviaQA, these answers belong to long and fluent sentences. 
CMRC-2018 is a dataset for Chinese Machine Reading Comprehension. It is a span-extraction reading comprehension dataset, which demands answers completely extracted from the original text. Following~\cite{dua-etal-2023-adapt}, we processed reading comprehension datasets (CMRC and MSMARCO) by shuffling their reference, aligning the formal definition of the OpenQA, where knowledge must be searched from the knowledge base, not given.

\subsection{Baselines}
To comprehensively compare and demonstrate the effectiveness of GenKI, we have chosen three different baselines. Task-specific baselines are the most competitive baselines on each dataset differently to make a fair comparison because different from GenKI, as these task-specific methods can not achieve consistent performance across all datasets. The latest baselines are essentially based on LLMs, as methods relying on LLMs consistently outperform traditional approaches. LLM backbones are the raw model used in our structure, demonstrating our enhancement to the basic model. The detailed baseline settings will be illustrated as follows:


\begin{table}[]
    \centering
    \caption{The detailed description and statistics of datasets.}
    \vspace{4mm}
    \label{tab:Datasetdescribe}
    \begin{tabular}{cccc}
    \toprule
        \textbf{Dataset} & \textbf{Format}&\textbf{Answer type} &  \textbf{ QA pairs}   \\
    \midrule 
        TriviaQA  &Free       &Entity &7993\\
        MSMARCO     &Free       & Sentence&8000   \\
        CMRC-2018   &Span      & Entity/Sentence &3219\\
    \bottomrule
    \end{tabular}
\end{table}
\subsubsection{TriviaQA specific baselines}

The baselines we choose to compare with our model in TriviaQA is the top three methods\footnote{https://paperswithcode.com/sota/question-answering-on-triviaqa} that share a similar scale with ours. 
This is because comparing our model with methods having a significantly larger scale of parameters would be unfair, as they require more time and computational resources for training than we do. These three methods include:
 \begin{itemize}
     \item \textbf{ChatGPT} leverages a transformer-based neural network to understand and generate human language. We choose the gpt-3.5-turbo-0301 version as a baseline for its stable performance.
     \item \textbf{Codex+REPLUG}~\cite{shi2023replug} employs retrieval systems to fetch relevant documents as prompt.
     \item \textbf{GLaM-62B}~\cite{du2022glam} uses a sparsely activated mixture-of-experts architecture to parallelize model.
 \end{itemize}
 

\subsubsection{MSMARCO specific baselines}
We select the top-3 models on MSMARCO question answering natural language generation task\footnote{https://microsoft.github.io/msmarco/} as baselines. These three methods include:
\begin{itemize}
    \item \textbf{PALM}~\cite{bi2020palm} alleviates the mismatch between pre-training and fine-tuning where generation is more than reconstructing the original text.
    \item \textbf{Masque NLGEN Style}~\cite{nishida2019multistyle} propose a multi-style summarization model regarding open domain question answering as summarizing on question and reference.
    \item \textbf{REAG} is an anonymous submit and gains third place. 
\end{itemize}

\subsubsection{CMRC specific baselines}
We also select the top-performing 3 models on CMRC-2018\footnote{https://paperswithcode.com/sota/chinese-reading-comprehension-on-cmrc-2018} as baselines, which include:
\begin{itemize}
    \item \textbf{ERNIE-Gram}~\cite{xiao-etal-2021-ernie}, an explicit n-gram masking method instead of token masking training.
    \item \textbf{MacBERT}~\cite{cui-etal-2020-revisiting} utilizes MLM task to train and mitigate the discrepancy with bidirectional encoder representations from transformers.
    \item \textbf{ERNIE2.0}~\cite{Sun_Wang_Li_Feng_Tian_Wu_Wang_2020} captures lexical, syntactic and semantic information in the training data.
\end{itemize}

\subsubsection{Latest baselines}
Despite these top-performing methods, we also adopt four latest LLM-based structures on OpenQA as baselines, following~\cite{gao2023retrieval} we classify them into retriever-augmented baselines (the first two) and generator-augmented baselines (the last two). We provide the same in-context learning prompt for all LLM-based baselines to ensure fairness:
\begin{itemize}
\vspace{-2mm}
    \item \textbf{LLM-KB}~\cite{ren2023investigating} uses LLM itself to augment the reference to enhance the zero-shot learning effect.
\vspace{-2mm}
    \item \textbf{LLaMA2-index}~\footnote{https://www.llamaindex.ai}, an end-to-end alignment and rerank structure to merge align retriever with LLM, which benefit from context augmentation.
\vspace{-2mm}
    \item \textbf{RFiD}~\cite{wang2023rfid} finetunes the decoder to generate answers by to differentiate between relationships and features.
    \item \textbf{Self-RAG}~\cite{asai2023self} fine-tuned an LLM generates, reflects, and critiques retrieved passages and their generations using special tokens.
\end{itemize}

\subsubsection{LLM backbone}
In order to further demonstrate the improvement of our framework GenKI, we also adopt the LLM backbones as baselines.
We use LLaMA-65B and GLM-6B as pre-trained LLM backbones, which allows us to explore different-scale models and also test the effectiveness of our framework from both perspectives of English and Chinese.


\subsection{Evaluation metrics}
Different datasets employ various evaluation methods for assessing answers.
For datasets like TriviaQA and CMRC-2018, which emphasize consistency with the answers, we use the Exact Match (EM), F1 score, and Recall as evaluation metrics. Specifically, in this experiment, the dataset consists of multiple answers, each of which is considered a correct answer. Therefore, the Recall and F1 metrics in this experiment differ from those of multiple-choice questions and are instead based on text-level Recall and F1. For example, if the correct answer is "large language model," and the generated answer is "language model," the Recall value would be 2/3. The final score for a question is determined by taking the highest value among all answers. To illustrate that our framework GenKI is capable of enabling models to generate fluent, extended content, we utilize BLEU and ROUGE metrics to evaluate the quality of answers in the MSMARCO dataset. Moreover, to gauge the fluency of model outputs, we employ the Coherence values derived from CTRLEval\cite{ke2022ctrleval}.  
We categorize these metrics into two categories, K and C, where metrics in the K class assess the model's knowledge integration capability, and metrics in the C class measure the model's controllable generation ability.

Among them, EM and F1 in TriviaQA and CMRC, BLEU and ROUGE in MSMARCO are in category K, while EM in CMRC, EM in TriviaQA and Coherence value in MSMARCO are in category C.

\subsection{Implementation details}
The instruction-tuning and model inference are conducted on 2 Tesla A100 80G GPUs for GLM-6B and 8 Tesla A800 80G GPUs for LLaMA-65B. 
Across all methods, we finetune the model with LoRA\cite{hu2021lora} with Lora-rank 32 in GLM and 8 in LLaMA.
The model parameters are optimized using the Adam\cite{kingma2017adam} with a default learning rate of 1e-4. 
We trained LLMs using the LoRA method to obtain three sets of fine-tuned parameters, each proposed for implementing full knowledge integration, retrieved knowledge integration and controllable generation. The total parameters of our structure were about 1(frozen LLM backbone) + 3*6\%(LoRA tunable parameters), altogether 7.08B in GLM and 76.7B in LLaMA.
Furthermore, to ensure fair comparisons, we tuned the parameters of all baseline models to their best performance. All the generator-augmented baselines (finetuned-LLM baselines) enjoy the same type of instruction tuning.

%% file: chapter6_Results.tex
\section{Results and Analysis}

\begin{table}[]
    \centering
    \caption{OpenQA performance of EM and F1-value on TriviaQA dataset.}
    \vspace{4mm}
    \label{tab:TriviaQA_datasets_benchmark}
    \begin{tabular}{cc|c|cc}
    \toprule
\textbf{Baseline type}& &\textbf{Method}  & \textbf{EM\%}    & \textbf{F1\%}    \\
    \midrule 
\multirow{3}{*}{\makecell{Task\\ specific \\ baselines}}&&GLaM-62B  & 75.8 & -       \\
&&Codex+REPLUG   & 77.3 & -     \\
&&Gpt-3.5-turbo-0301    & 77.9  & 83.9  \\
    \midrule 
\multirow{4}{*}{\makecell{Latest \\ baselines}}&\multirow{2}{*}{\makecell{RA}}&LLM-KB   & 74.8 & 80.1 \\ 
&&LLaMA2-index &67.4\footnotemark{} &-\\
&\multirow{2}{*}{\makecell{GA}}&RFiD      & 72.7 & - \\
&&Self-RAG      & 69.3 & -\\
    \midrule 
\multirow{2}{*}{\makecell{LLM \\ backbones }}&&GLM-6B$_{B}$ & 7.3 & 20.8       \\
&&LLaMA-65B$_{B}$ & 77.8  & 82.3 \\
    \midrule 
\multirow{2}{*}{Our work}&&GLM-6B$_{G}$  & 72.4 & 78.6       \\
&&LLaMA-65B$_{G}$   & \textbf{81.6}  & \textbf{86.9} \\
      \bottomrule
    \end{tabular}
    
    \begin{tablenotes}
        \footnotesize
        \item[*] $RA$: retriever-augmented baselines, $GA$: generator-augmented baselines (finetuned-LLM baselines), $B$: base model, $G$: GenKI
    \end{tablenotes}
    
\end{table}
\footnotetext{We employ the recall metric here to enhance the performance of llaMA2-index. This choice is made due to the inadequacy of assessing a long-form answering method in a short-form answering dataset using the EM-score.} 
In this section, we first present a performance comparison between our method and the baseline across three datasets. Then, through a detailed analysis, we demonstrate how the two main modules in GenKI, i.e., knowledge integration and controllable generation, work for OpenQA on LLMs.

\subsection{Overall performance}
We present the overall performance of GenKI and all baselines in tables~\ref{tab:TriviaQA_datasets_benchmark}, ~\ref{tab:MSMARCO datasets benchmark} and ~\ref{tab:CMRC datasets benchmark} on TriviaQA, MSMARCO, and CMRC, respectively.

In \textbf{TriviaQA}, the LLaMA model, optimized with our framework, successfully achieved a 4.8\% gain on the EM metric and a 5.6\% gain on the F1 metric compared to the backbone itself. 
The GLM model, optimized with our framework, successfully gained 891.7\% and 277.8\% on the EM and F1 metrics, respectively, compared to the backbone itself. Specifically, the LLaMA model is compared against the current retrieval-augmented (REPLUG, LLM-KB, LLaMA2-index), generator-augmented (RFiD, Self-RAG) and few-shot learning (ChatGPT, GLaM-62B) approaches in OpenQA, outperforming them by 4.7\% and 3.6\% on the EM and F1 metrics, respectively. Moreover, it achieved better performance to ChatGPT, which is currently regarded as one of the best LLMs.

\begin{table}[t!]
    \centering
    \caption{OpenQA performance of BLEU and ROUGE on MSMARCO dataset.}
    \vspace{4mm}
    \label{tab:MSMARCO datasets benchmark}
    \begin{tabular}{C{1.2cm}c|c|C{1.2cm}C{1.5cm}}
    \toprule
\textbf{\makecell{Baseline\\type}}&&\textbf{Method}&\textbf{BLEU-1}&\textbf{ROUGE-L}\\
    \midrule 
\multirow{3}{*}{\makecell{Task  \\ specific\\baselines}}&&Masque NLGEN     & 0.501   & 0.496  \\
&&PALM & 0.499   & 0.498      \\
&&REAG    & 0.497   & \textbf{0.498}  \\
    \midrule 
\multirow{4}{*}{\makecell{Latest \\ baselines}}&\multirow{2}{*}{\makecell{RA}} &LLM-KB & 0.262   & 0.241      \\
&&LLaMA2-index & 0.588 & 0.447 \\
&\multirow{2}{*}{\makecell{GA}}&RFiD & 0.561   & 0.454\\
&&Self-RAG & 0.567   & 0.351      \\
    \midrule 
\multirow{2}{*}{\makecell{LLM  \\ backbones}}&&LLaMA-65B$_{B}$ & 0.473   & 0.278      \\
&&GLM-6B$_{B}$ & 0.475   & 0.248      \\

    \midrule
\multirow{2}{*}{Our work}&&LLaMA-65B$_{G}$ &0.521 &0.317\\
&&GLM-6B$_{G}$ & \textbf{0.598}   & 0.456      \\
      \bottomrule
    \end{tabular}
\end{table}


In the \textbf{MSMARCO} dataset, the GLM model, optimized with our framework, outperforms the backbone itself by 25.9\% and 83.8\% on the BLEU-1 and ROUGE-L metrics. However, LLaMA falls short compared to GLM in terms of performance. This could possibly be attributed to its lack of a supervised training process, which leads to difficulties in generating lengthy and coherent text. Even so, LLaMA still shows improvement by 10.1\% and 14.0\% on the BLEU-1 and ROUGE-L metrics.
Our GenKI framework outperforms the best model by a margin of 1.7\% on the BLEU-1 metric.
Nevertheless, in our framework, the ROUGE-L metric lagged behind the performance of the current top-performing approach. This discrepancy arises due to the fact that the references in the MSMARCO dataset comprise lengthier passages that closely resemble the source text, contrasting with knowledge integration capacity of our framework.

Within the \textbf{CMRC} dataset, the GLM model outperforms the backbone itself by a significant margin, exceeding the baseline performance by over 40 times. Meanwhile, the GLM model shows improvements over the existing state-of-the-art approach by 5.4\% and 0.9\% on the EM and F1 metrics, respectively. This accomplishment is particularly remarkable given that CMRC-2018 operates as a sequence labeling dataset. Generative models inherently encounter challenges in this context, as they must generate results that align precisely with the source text—a requirement that inherently presents a natural disadvantage. We refrained from employing the LLaMA and RFiD models on this dataset due to their inadequate performance in handling the Chinese language. To further analyze and evaluate the effectiveness of our model, we conducted the following analyses on two special cases described below:

\subsubsection{Can GenKI work when the entire knowledge base is inaccessible?}
\begin{figure}[t]
  \includegraphics[width=\linewidth]{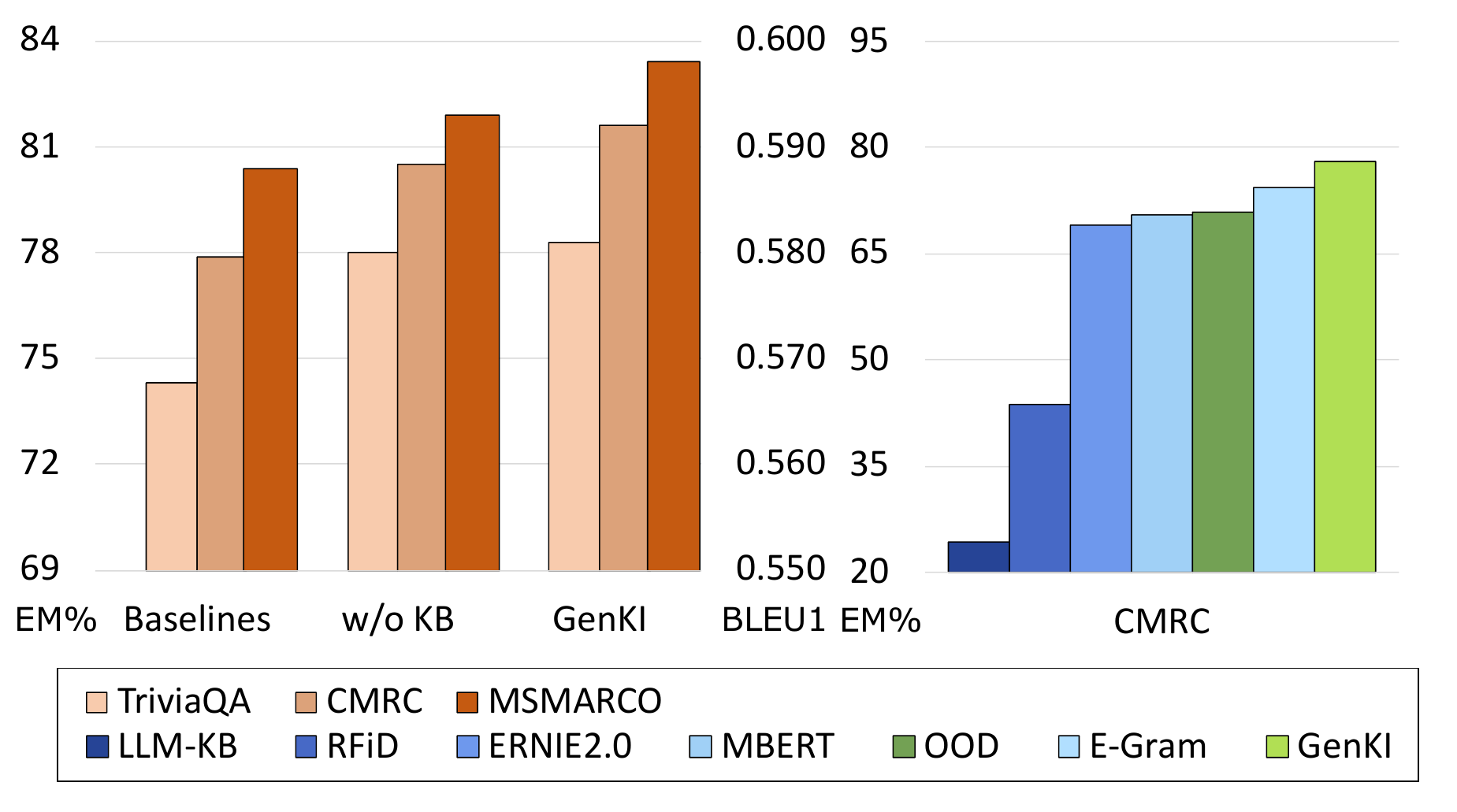}
  \caption{Comparison of (Above) GenKI, GenKI without knowledge base and best baseline, (Below) GenKI, GenKI in Out-Of-Domain scenario and baselines.}
  \label{fig:ood}
\end{figure}

In practical academic and industry contexts, models frequently depend on search engines to gather requisite knowledge, facing limitations in leveraging the entire knowledge base for fine-tuning. In such instances, our model ($X+K_R+\mathrm{PostP}$) consistently surpasses baseline performances, demonstrating superiority by 3.3\% in TriviaQA, 0.8\% in MSMARCO, and 4.9\% in CMRC. The comparative results are illustrated in Figure~\ref{fig:ood}.

\subsubsection{Can GenKI adapt to different formats of answers?}
To demonstrate the adaptability of our model to another format domain, we partition the CMRC dataset, encompassing both long and short-answer formats by their average length. Specifically, we fine-tune our model only on the short-answer segment during step $\mathrm{PostP}$, resulting in a noteworthy result of 71.70 on the EM metric. This achievement surpasses all recent baselines, trailing behind only one of the task-specific baselines. Notably, given that these task-specific baselines are exclusively trained intra-domain, our model exhibits exceptional performance, underscoring its remarkable out-of-domain capability, which is shown in Figure~\ref{fig:ood}.

To sum up, analysis of these three datasets demonstrates that GenKI has performed remarkable ability in learning retrieved knowledge and multi-format controllable generation compared with task-specific, latest baselines both on retriever and generator augmentation and LLM backbones.
\begin{table}[t!]
    \centering
    \caption{OpenQA performance of EM and F1-value on CMRC dataset.}
    \label{tab:CMRC datasets benchmark}
    \vspace{4mm}
    \begin{tabular}{cc|c|cc}
    \toprule
\textbf{Baseline type   }       && \textbf{Method}         & \textbf{EM\%}   & \textbf{F1\%} \\
    \midrule 
\multirow{3}{*}{\makecell{Task specific \\ baselines}}&&ERNIE2.0   & 69.1   & 88.6  \\
&&MacBERT   & 70.7   & 88.9  \\
&&ERNIE-Gram    & 74.3   & 90.5  \\
    \midrule 
\multirow{2}{*}{\makecell{Latest \\ baselines}}&RA&LLM-KB     & 24.3   & 48.8  \\
&GA&RFiD &43.7 &65.1\\
    \midrule 
LLM backbones&&GLM-6B$_{\mathrm{base}}$  & 1.8   &47.7      \\
    \midrule 
Our work&&GLM-6B$_{\mathrm{GenKI}}$ & \textbf{78.3}   &\textbf{91.1}      \\
      \bottomrule
    \end{tabular}
    \begin{tablenotes}
        \footnotesize
        \item[*] $RA$: retriever-augmented baselines, $GA$: generator-augmented baselines (finetuned-LLM baselines)
    \end{tablenotes}
\end{table}

\subsection{Analysis on knowledge integration module}

\subsubsection{To what extent can the acquired knowledge impact the proficiency of the model?}


In this section, we set aside all output format requirements and focus solely on studying the model's grasp and understanding of knowledge. The TriviaQA and CMRC datasets, which are more relevant in terms of knowledge, are chosen to facilitate the investigation of this ability better. Additionally, we adopt the weakest format requirement, recall value (where the model mentions words in the answer), for our study. As the ablation results are shown in table~\ref{tab:Complete_TC}, we observe an improvement of recall by 19.4\% and 6.2\% in TriviaQA for the GLM and LLaMA models, respectively, after applying $K_R$. 
To ascertain the link between this enhancement and the quality of the retrieval results, we conduct comparative experiments under varying retrieval quality levels. 

We measure the quality of retrieval results by utilizing the frequency of occurrence of the ground truth within the retrieval results. 
For instance, if the ground truth is ``Large Language Model" and our retrieval result is ``Large language models have gained widespread language applications." then the quality of the retrieval result would be calculated as 3/8 = 0.375, without calculating the latter occurrence of ``language". 
Subsequently, the effectiveness of the model's integration of knowledge is evaluated using the Recall metric. 
Additionally, we constrain the maximum output length of the model to prevent the model from achieving a high recall by producing excessively redundant results. 
We then select a subset of examples from TriviaQA, consisting of both higher-quality and lower-quality retrievals, to gain more data for studying the impact of retrieval quality on the recall values of the model's generated answer.

\begin{figure}[t]
  \includegraphics[width=\linewidth]{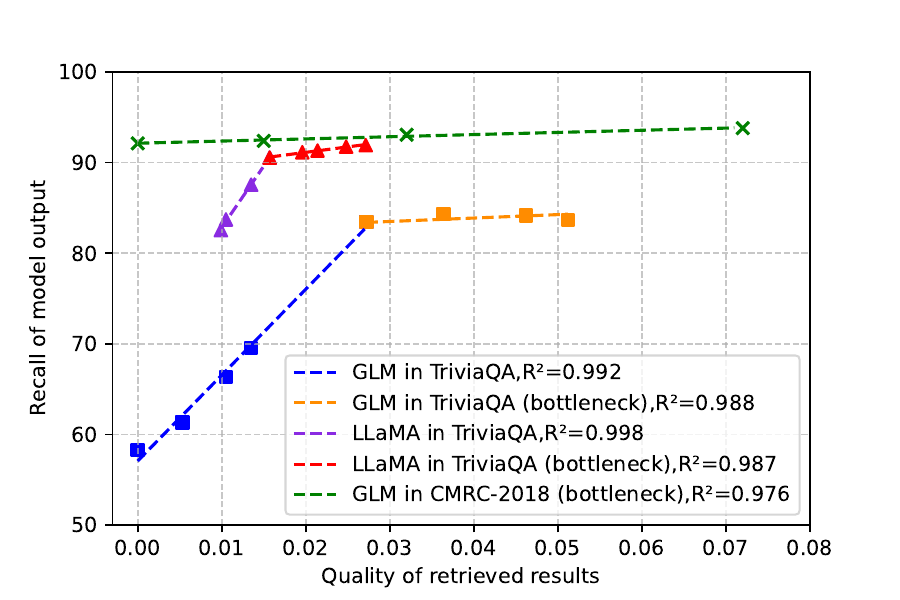}
  \caption{The linear fitting relationship between quality of retrieved result and knowledge integration effect}
  \label{fig:retrieve_recall_impact}
\end{figure}

The relationship is demonstrated in Figure~\ref{fig:retrieve_recall_impact}. From this figure, we can infer two pieces of information:

(1) The relationship between retrieval quality and model recall initially exhibited a linear trend across these examples ($R^2 \textgreater$ 0.99), but eventually reached a bottleneck state, transitioning into another linear relationship ($R^2 \textgreater$ 0.985).
This suggests that the model has been extensively integrated with the necessary knowledge. The ultimate bottleneck of the model is correlated with the scale of the model. For instance, the bottleneck for LLaMA-65B occurred at around 90\% recall, while the bottleneck for GLM-6B was observed at around 80\% recall.

\begin{table*}[t]
    \centering
    \caption{Ablation experimental results on TriviaQA.}
    \vspace{4mm}
    \label{tab:Complete_TC}
    
\begin{threeparttable}
    \begin{tabular}{lccclccc}
    \toprule
$X$=GLM-6B          & EM\%    & F1\%    & Recall\% & $X$=LLaMa-65B   & EM\%    & F1\%    & Recall\%  \\ 
  \midrule 
$X$       & 7.3  & 20.8 & 58.3              & $X$       & 77.8 & 82.3 & 82.5           \\
$X+K$     & 47.4 & 61.7 & 66.3              & $X+K$     & 72.7 & 79.0 & 83.7            \\
$X+K+\mathrm{PostP}$   & 57.1 & 64.7 & 64.9              &  $X+K+\mathrm{PostP}$  & 75.9 & 80.9 & 82.8        \\
$X+K_R$   & 54.0 & 66.2 & 69.6   & $X+K_R$   & 72.7 & 80.1 & 87.6  \\
$X+K_R+\mathrm{PostP}$ & 58.0 & 66.6 & 67.2              & $X+K_R+\mathrm{PostP}$ & 80.5 & 85.4 & 86.9 \\
Rm      & 58.6   & 68.2 & 69.0 & Rm & 80.3 & 85.4 & 85.6              \\
Ext      & 72.4   & 78.6 & 79.8 & Ext & 81.5 & 86.5 & 88.7           \\
GenKI      & 72.4   & 78.6 & 79.9 & GenKI & 81.6 & 86.7 & 87.2           \\
      \bottomrule
    \end{tabular}
    \begin{tablenotes}
        \footnotesize
        \item[*] $X$: LLM backbone, $K$:output after full knowledge integration model, $K_R$: output after retrieved knowledge integration model
        \item[**] $\mathrm{PostP}$: output after controllable generation, Rm: output after reward model, Ext: external model selection
    \end{tablenotes}
\end{threeparttable}
\end{table*}

\begin{table*}[h]
    \centering
    \caption{Ablation experimental results on MSMARCO and CMRC.}
    \vspace{4mm}
    \label{tab:Complete_M}
    
\begin{threeparttable}
    \begin{tabular}{lcccc|lccc}
    \toprule
    
    \multicolumn{5}{c|}{MSMARCO} &  \multicolumn{4}{c}{CMRC-2018} \\
    \midrule
$X$=GLM-6B    & BLEU-1& BLEU-2 & ROUGE-L & Coherence & $X$=GLM-6B          & EM\%    & F1\%    & Recall\%\\
    \midrule 
$X$ & 0.475 & 0.387&0.248&-4.164   & $X$      & 1.8  & 47.7 & 92.1   \\
$X+K$    & 0.588  & 0.523   & 0.428   & -4.195   & $X+K$    & 33.5 & 71.8 & 93.1   \\
$X+K+\mathrm{PostP}$   & 0.589 & 0.526  & 0.455   & -4.185       & $X+K+\mathrm{PostP}$ & 77.8 & 91.0 & 92.3   \\
$X+K_R$   & 0.594 & 0.526 & 0.429   & -4.196  & $X+K_R$  & 37.1 & 73.3 & 93.8   \\
$X+K_R+\mathrm{PostP}$       & 0.593 & 0.529  & 0.452   & -4.198   & $X+K_R+\mathrm{PostP}$ &  78.0     &  91.1     &   92.2 \\
Rm      & 0.597 & 0.531 & 0.457   & -4.184 & Rm &  78.1    &  91.3     &  92.8   \\
Ext      & 0.598  & 0.534& 0.457   & -4.195  & Ext &  75.2    &  89.9     &  91.4    \\
GenKI       & 0.598 & 0.534   & 0.456   & -4.174    & GenKI &  78.3    &  91.1     &  92.3    \\
      \bottomrule
    \end{tabular}
        \begin{tablenotes}
        \footnotesize
        \item[*] Coherence in CTRLEval
    \end{tablenotes}
\end{threeparttable}
\end{table*}

(2) By comparing the recall of the LLM backbone with the recall of the LLM integrated with full knowledge($X+K$ on table~\ref{tab:Complete_TC},~\ref{tab:Complete_M}), or whether the model reaches a bottleneck, we can infer whether the pretraining of the model includes the required knowledge of target dataset.

First, it can be inferred that the GLM's pre-training hardly contains Wikipedia knowledge (58.3 compared to 66.3), but found a bottleneck in the CMRC from the beginning. This observation aligns with our expectations, given that GLM focuses more on Chinese language corpora. Additionally, it can be inferred that LLaMA's pretraining contains knowledge from Wikidata(82.5 similar to 83.7), which is also supported by findings in LLaMA's technical report ~\cite{touvron2023llama}.

However, merely enhancing the model's knowledge understanding through the Knowledge integration module is insufficient. Overemphasizing the model's knowledge understanding may weaken its controllable generation ability. Specifically, the performance of both $X+K$ and $X+K_R$ is better than $X$ when using GLM-6B but worse when using LLaMa-65B. This is also why our work designs a novel three-stage paradigm of retrieval, knowledge integration, and controllable generation, avoiding the distributional differences of LLM in the two-stage RAG. Next, this paper will analyze the controllable generation module.

\subsection{Analysis on controllable generation module}
\subsubsection{Can GenKI tackle distribution between knowledge integration or controlled generation?}
The success of our model lies in effectively balancing both knowledge integration and controlled generation. To substantiate this claim, we categorize the metrics into two components: knowledge integration metric $K_m$ (EM$\%$, F1$\%$, Recall$\%$, BLEU-1, BLEU-2 and Rouge-L) and controlled generation metric $C_m$ (EM$\%$, BLEU-1, BLEU-2, Rouge-L and Coherence). As shown in table~\ref{tab:Complete_TC} and table~\ref{tab:Complete_M} our model succeeds in obtaining all $K_m\&C_m$ and $C_m$ metrics and most $K_m$ metrics. 

We are also concerned about how different structures influence our answers. 
After undergoing Step $\mathrm{PostP}$ in table~\ref{tab:Complete_TC}, the recall of all answers decreases (by up to 2\%), while the EM of all answers increases(by up to 2 times). This indicates that the model is more inclined to have learned distribution of instructions which is different from knowledge. This observation underscores the importance of splitting the answer generation process into two steps: knowledge integration and controllable generation. The $\mathrm{Ext}$ step plays the same effect on the recall of answers. It performs better when the backbone LLM performs worse. The improvement reaches 18.8\% recall in TriviaQA using GLM-6B.

The reward model step, on the other hand, demonstrates improvements in coherence(by 1.8\%) within the MSMARCO dataset, as shown in table~\ref{tab:Complete_M}. Our reward model module achieved better results in all datasets, except it slightly loses efficacy on LLaMA in TriviaQA, which aims to significantly enhance both answer coherence and consistency to better align with common human habits and preferences.

The introduction of the external model selection module has achieved significant success on knowledge question-answering datasets such as CMRC and TriviaQA. Particularly, for models with weaker knowledge mastery like GLM-6B, there is a substantial improvement (24.8$\%$ in EM-score). This indicates that the introduction of this module can effectively ensure the accuracy of model knowledge. Additionally, our data also confirms its compatibility with our framework.

%% file: chapter7_Conclusion.tex
\section{Conclusion}


In this paper, we introduced GenKI, a novel OpenQA framework that combines knowledge retrieval, knowledge integration, and controllable generation. By splitting the generation process of LLMs into two distinct phases of knowledge integration and controllable generation, we addressed the challenges of knowledge integration deficiency and answer format misalignment. GenKI has demonstrated enhanced performance in OpenQA task, surpassing both traditional methods and other LLM-based approaches. This research not only explores the relationship between provided search results and knowledge generation within LLMs but also propels LLM-based OpenQA towards the desired format. 


%% file: main.bbl
\begin{thebibliography}{10}

\bibitem{abdallah2023generator}
Abdelrahman Abdallah and Adam Jatowt.
\newblock Generator-retriever-generator: A novel approach to open-domain question answering.
\newblock {\em arXiv preprint arXiv:2307.11278}, 2023.

\bibitem{aizawa2003information}
Akiko Aizawa.
\newblock An information-theoretic perspective of tf--idf measures.
\newblock {\em Information Processing \& Management}, 39(1):45--65, 2003.

\bibitem{asai2023self}
Akari Asai, Zeqiu Wu, Yizhong Wang, Avirup Sil, and Hannaneh Hajishirzi.
\newblock Self-rag: Learning to retrieve, generate, and critique through self-reflection.
\newblock {\em arXiv preprint arXiv:2310.11511}, 2023.

\bibitem{bajaj2018ms}
Payal Bajaj, Daniel Campos, Nick Craswell, Li~Deng, Jianfeng Gao, Xiaodong Liu, Rangan Majumder, Andrew McNamara, Bhaskar Mitra, Tri Nguyen, Mir Rosenberg, Xia Song, Alina Stoica, Saurabh Tiwary, and Tong Wang.
\newblock Ms marco: A human generated machine reading comprehension dataset, 2018.

\bibitem{DBLP:conf/nips/BevilacquaOLY0P22}
Michele Bevilacqua, Giuseppe Ottaviano, Patrick S.~H. Lewis, Scott Yih, Sebastian Riedel, and Fabio Petroni.
\newblock Autoregressive search engines: Generating substrings as document identifiers.
\newblock In {\em NeurIPS}, 2022.

\bibitem{bi2020palm}
Bin Bi, Chenliang Li, Chen Wu, Ming Yan, Wei Wang, Songfang Huang, Fei Huang, and Luo Si.
\newblock Palm: Pre-training an autoencoding autoregressive language model for context-conditioned generation, 2020.

\bibitem{brown2020language}
Tom Brown, Benjamin Mann, Nick Ryder, Melanie Subbiah, Jared~D Kaplan, Prafulla Dhariwal, Arvind Neelakantan, Pranav Shyam, Girish Sastry, Amanda Askell, et~al.
\newblock Language models are few-shot learners.
\newblock {\em Advances in neural information processing systems}, 33:1877--1901, 2020.

\bibitem{chowdhery2022palm}
Aakanksha Chowdhery, Sharan Narang, Jacob Devlin, Maarten Bosma, Gaurav Mishra, Adam Roberts, Paul Barham, Hyung~Won Chung, Charles Sutton, Sebastian Gehrmann, et~al.
\newblock Palm: Scaling language modeling with pathways.
\newblock {\em arXiv preprint arXiv:2204.02311}, 2022.

\bibitem{DBLP:journals/corr/abs-2204-02311}
Aakanksha Chowdhery, Sharan Narang, Jacob Devlin, Maarten Bosma, Gaurav Mishra, Adam Roberts, Paul Barham, Hyung~Won Chung, Charles Sutton, Sebastian Gehrmann, Parker Schuh, Kensen Shi, Sasha Tsvyashchenko, Joshua Maynez, Abhishek Rao, Parker Barnes, Yi~Tay, Noam Shazeer, Vinodkumar Prabhakaran, Emily Reif, Nan Du, Ben Hutchinson, Reiner Pope, James Bradbury, Jacob Austin, Michael Isard, Guy Gur{-}Ari, Pengcheng Yin, Toju Duke, Anselm Levskaya, Sanjay Ghemawat, Sunipa Dev, Henryk Michalewski, Xavier Garcia, Vedant Misra, Kevin Robinson, Liam Fedus, Denny Zhou, Daphne Ippolito, David Luan, Hyeontaek Lim, Barret Zoph, Alexander Spiridonov, Ryan Sepassi, David Dohan, Shivani Agrawal, Mark Omernick, Andrew~M. Dai, Thanumalayan~Sankaranarayana Pillai, Marie Pellat, Aitor Lewkowycz, Erica Moreira, Rewon Child, Oleksandr Polozov, Katherine Lee, Zongwei Zhou, Xuezhi Wang, Brennan Saeta, Mark Diaz, Orhan Firat, Michele Catasta, Jason Wei, Kathy Meier{-}Hellstern, Douglas Eck, Jeff Dean, Slav Petrov, and Noah Fiedel.
\newblock Palm: Scaling language modeling with pathways.
\newblock {\em CoRR}, abs/2204.02311, 2022.

\bibitem{cui-etal-2020-revisiting}
Yiming Cui, Wanxiang Che, Ting Liu, Bing Qin, Shijin Wang, and Guoping Hu.
\newblock Revisiting pre-trained models for {C}hinese natural language processing.
\newblock In {\em Findings of the Association for Computational Linguistics: EMNLP 2020}, pages 657--668, Online, November 2020. Association for Computational Linguistics.

\bibitem{cui-emnlp2019-cmrc2018}
Yiming Cui, Ting Liu, Wanxiang Che, Li~Xiao, Zhipeng Chen, Wentao Ma, Shijin Wang, and Guoping Hu.
\newblock A span-extraction dataset for {C}hinese machine reading comprehension.
\newblock In {\em Proceedings of the 2019 Conference on Empirical Methods in Natural Language Processing and the 9th International Joint Conference on Natural Language Processing (EMNLP-IJCNLP)}, pages 5886--5891, Hong Kong, China, November 2019. Association for Computational Linguistics.

\bibitem{DBLP:conf/iclr/DasDZM19}
Rajarshi Das, Shehzaad Dhuliawala, Manzil Zaheer, and Andrew McCallum.
\newblock Multi-step retriever-reader interaction for scalable open-domain question answering.
\newblock In {\em 7th International Conference on Learning Representations, {ICLR} 2019, New Orleans, LA, USA, May 6-9, 2019}. OpenReview.net, 2019.

\bibitem{devlin2018bert}
Jacob Devlin, Ming-Wei Chang, Kenton Lee, and Kristina Toutanova.
\newblock Bert: Pre-training of deep bidirectional transformers for language understanding.
\newblock {\em arXiv preprint arXiv:1810.04805}, 2018.

\bibitem{du2022glam}
Nan Du, Yanping Huang, Andrew~M Dai, Simon Tong, Dmitry Lepikhin, Yuanzhong Xu, Maxim Krikun, Yanqi Zhou, Adams~Wei Yu, Orhan Firat, et~al.
\newblock Glam: Efficient scaling of language models with mixture-of-experts.
\newblock In {\em International Conference on Machine Learning}, pages 5547--5569. PMLR, 2022.

\bibitem{du2021glm}
Zhengxiao Du, Yujie Qian, Xiao Liu, Ming Ding, Jiezhong Qiu, Zhilin Yang, and Jie Tang.
\newblock Glm: General language model pretraining with autoregressive blank infilling.
\newblock {\em arXiv preprint arXiv:2103.10360}, 2021.

\bibitem{dua-etal-2023-adapt}
Dheeru Dua, Emma Strubell, Sameer Singh, and Pat Verga.
\newblock To adapt or to annotate: Challenges and interventions for domain adaptation in open-domain question answering.
\newblock In Anna Rogers, Jordan Boyd-Graber, and Naoaki Okazaki, editors, {\em Proceedings of the 61st Annual Meeting of the Association for Computational Linguistics (Volume 1: Long Papers)}, pages 14429--14446, Toronto, Canada, July 2023. Association for Computational Linguistics.

\bibitem{gao2023retrieval}
Yunfan Gao, Yun Xiong, Xinyu Gao, Kangxiang Jia, Jinliu Pan, Yuxi Bi, Yi~Dai, Jiawei Sun, and Haofen Wang.
\newblock Retrieval-augmented generation for large language models: A survey.
\newblock {\em arXiv preprint arXiv:2312.10997}, 2023.

\bibitem{gu2025rapid}
Hongchao Gu, Dexun Li, Kuicai Dong, Hao Zhang, Hang Lv, Hao Wang, Defu Lian, Yong Liu, and Enhong Chen.
\newblock Rapid: Efficient retrieval-augmented long text generation with writing planning and information discovery.
\newblock {\em arXiv preprint arXiv:2503.00751}, 2025.

\bibitem{hoffmann2022training}
Jordan Hoffmann, Sebastian Borgeaud, Arthur Mensch, Elena Buchatskaya, Trevor Cai, Eliza Rutherford, Diego de~Las Casas, Lisa~Anne Hendricks, Johannes Welbl, Aidan Clark, et~al.
\newblock Training compute-optimal large language models.
\newblock {\em arXiv preprint arXiv:2203.15556}, 2022.

\bibitem{hu2021lora}
Edward~J Hu, Yelong Shen, Phillip Wallis, Zeyuan Allen-Zhu, Yuanzhi Li, Shean Wang, Lu~Wang, and Weizhu Chen.
\newblock Lora: Low-rank adaptation of large language models.
\newblock {\em arXiv preprint arXiv:2106.09685}, 2021.

\bibitem{DBLP:journals/corr/abs-2007-01282}
Gautier Izacard and Edouard Grave.
\newblock Leveraging passage retrieval with generative models for open domain question answering.
\newblock {\em CoRR}, abs/2007.01282, 2020.

\bibitem{joshi2020spanbert}
Mandar Joshi, Danqi Chen, Yinhan Liu, Daniel~S Weld, Luke Zettlemoyer, and Omer Levy.
\newblock Spanbert: Improving pre-training by representing and predicting spans.
\newblock {\em Transactions of the association for computational linguistics}, 8:64--77, 2020.

\bibitem{joshi-etal-2017-triviaqa}
Mandar Joshi, Eunsol Choi, Daniel Weld, and Luke Zettlemoyer.
\newblock {T}rivia{QA}: A large scale distantly supervised challenge dataset for reading comprehension.
\newblock In {\em Proceedings of the 55th Annual Meeting of the Association for Computational Linguistics (Volume 1: Long Papers)}, pages 1601--1611, Vancouver, Canada, July 2017. Association for Computational Linguistics.

\bibitem{kandpal2023large}
Nikhil Kandpal, Haikang Deng, Adam Roberts, Eric Wallace, and Colin Raffel.
\newblock Large language models struggle to learn long-tail knowledge.
\newblock In {\em International Conference on Machine Learning}, pages 15696--15707. PMLR, 2023.

\bibitem{karpukhin-etal-2020-dense}
Vladimir Karpukhin, Barlas Oguz, Sewon Min, Patrick Lewis, Ledell Wu, Sergey Edunov, Danqi Chen, and Wen-tau Yih.
\newblock Dense passage retrieval for open-domain question answering.
\newblock In {\em Proceedings of the 2020 Conference on Empirical Methods in Natural Language Processing (EMNLP)}, pages 6769--6781, Online, November 2020. Association for Computational Linguistics.

\bibitem{ke-etal-2022-ctrleval}
Pei Ke, Hao Zhou, Yankai Lin, Peng Li, Jie Zhou, Xiaoyan Zhu, and Minlie Huang.
\newblock {CTRLE}val: An unsupervised reference-free metric for evaluating controlled text generation.
\newblock In Smaranda Muresan, Preslav Nakov, and Aline Villavicencio, editors, {\em Proceedings of the 60th Annual Meeting of the Association for Computational Linguistics (Volume 1: Long Papers)}, pages 2306--2319, Dublin, Ireland, May 2022. Association for Computational Linguistics.

\bibitem{ke2022ctrleval}
Pei Ke, Hao Zhou, Yankai Lin, Peng Li, Jie Zhou, Xiaoyan Zhu, and Minlie Huang.
\newblock Ctrleval: An unsupervised reference-free metric for evaluating controlled text generation, 2022.

\bibitem{kingma2017adam}
Diederik~P. Kingma and Jimmy Ba.
\newblock Adam: A method for stochastic optimization, 2017.

\bibitem{DBLP:conf/nips/LewisPPPKGKLYR020}
Patrick S.~H. Lewis, Ethan Perez, Aleksandra Piktus, Fabio Petroni, Vladimir Karpukhin, Naman Goyal, Heinrich K{\"{u}}ttler, Mike Lewis, Wen{-}tau Yih, Tim Rockt{\"{a}}schel, Sebastian Riedel, and Douwe Kiela.
\newblock Retrieval-augmented generation for knowledge-intensive {NLP} tasks.
\newblock In Hugo Larochelle, Marc'Aurelio Ranzato, Raia Hadsell, Maria{-}Florina Balcan, and Hsuan{-}Tien Lin, editors, {\em Advances in Neural Information Processing Systems 33: Annual Conference on Neural Information Processing Systems 2020, NeurIPS 2020, December 6-12, 2020, virtual}, 2020.

\bibitem{li2023selfprompting}
Junlong Li, Zhuosheng Zhang, and Hai Zhao.
\newblock Self-prompting large language models for zero-shot open-domain qa, 2023.

\bibitem{liu2019roberta}
Yinhan Liu, Myle Ott, Naman Goyal, Jingfei Du, Mandar Joshi, Danqi Chen, Omer Levy, Mike Lewis, Luke Zettlemoyer, and Veselin Stoyanov.
\newblock Roberta: A robustly optimized bert pretraining approach.
\newblock {\em arXiv preprint arXiv:1907.11692}, 2019.

\bibitem{ma2023query}
Xinbei Ma, Yeyun Gong, Pengcheng He, Hai Zhao, and Nan Duan.
\newblock Query rewriting for retrieval-augmented large language models.
\newblock {\em arXiv preprint arXiv:2305.14283}, 2023.

\bibitem{DBLP:conf/acl/MallenAZDKH23}
Alex Mallen, Akari Asai, Victor Zhong, Rajarshi Das, Daniel Khashabi, and Hannaneh Hajishirzi.
\newblock When not to trust language models: Investigating effectiveness of parametric and non-parametric memories.
\newblock In Anna Rogers, Jordan~L. Boyd{-}Graber, and Naoaki Okazaki, editors, {\em Proceedings of the 61st Annual Meeting of the Association for Computational Linguistics (Volume 1: Long Papers), {ACL} 2023, Toronto, Canada, July 9-14, 2023}, pages 9802--9822. Association for Computational Linguistics, 2023.

\bibitem{mao-etal-2021-generation}
Yuning Mao, Pengcheng He, Xiaodong Liu, Yelong Shen, Jianfeng Gao, Jiawei Han, and Weizhu Chen.
\newblock Generation-augmented retrieval for open-domain question answering.
\newblock In Chengqing Zong, Fei Xia, Wenjie Li, and Roberto Navigli, editors, {\em Proceedings of the 59th Annual Meeting of the Association for Computational Linguistics and the 11th International Joint Conference on Natural Language Processing (Volume 1: Long Papers)}, pages 4089--4100, Online, August 2021. Association for Computational Linguistics.

\bibitem{min2021recent}
Bonan Min, Hayley Ross, Elior Sulem, Amir Pouran~Ben Veyseh, Thien~Huu Nguyen, Oscar Sainz, Eneko Agirre, Ilana Heintz, and Dan Roth.
\newblock Recent advances in natural language processing via large pre-trained language models: A survey.
\newblock {\em ACM Computing Surveys}, 2021.

\bibitem{DBLP:journals/corr/abs-1911-03868}
Sewon Min, Danqi Chen, Luke Zettlemoyer, and Hannaneh Hajishirzi.
\newblock Knowledge guided text retrieval and reading for open domain question answering.
\newblock {\em CoRR}, abs/1911.03868, 2019.

\bibitem{nguyen2023cof}
Hoang~H Nguyen, Ye~Liu, Chenwei Zhang, Tao Zhang, and Philip~S Yu.
\newblock Cof-cot: Enhancing large language models with coarse-to-fine chain-of-thought prompting for multi-domain nlu tasks.
\newblock {\em arXiv preprint arXiv:2310.14623}, 2023.

\bibitem{nishida2019multistyle}
Kyosuke Nishida, Itsumi Saito, Kosuke Nishida, Kazutoshi Shinoda, Atsushi Otsuka, Hisako Asano, and Junji Tomita.
\newblock Multi-style generative reading comprehension, 2019.

\bibitem{ouyang2022training}
Long Ouyang, Jeffrey Wu, Xu~Jiang, Diogo Almeida, Carroll Wainwright, Pamela Mishkin, Chong Zhang, Sandhini Agarwal, Katarina Slama, Alex Ray, et~al.
\newblock Training language models to follow instructions with human feedback.
\newblock {\em Advances in Neural Information Processing Systems}, 35:27730--27744, 2022.

\bibitem{petroni-etal-2019-language}
Fabio Petroni, Tim Rockt{\"a}schel, Sebastian Riedel, Patrick Lewis, Anton Bakhtin, Yuxiang Wu, and Alexander Miller.
\newblock Language models as knowledge bases?
\newblock In {\em Proceedings of the 2019 Conference on Empirical Methods in Natural Language Processing and the 9th International Joint Conference on Natural Language Processing (EMNLP-IJCNLP)}, pages 2463--2473, Hong Kong, China, November 2019. Association for Computational Linguistics.

\bibitem{ren2023investigating}
Ruiyang Ren, Yuhao Wang, Yingqi Qu, Wayne~Xin Zhao, Jing Liu, Hao Tian, Hua Wu, Ji-Rong Wen, and Haifeng Wang.
\newblock Investigating the factual knowledge boundary of large language models with retrieval augmentation.
\newblock {\em arXiv preprint arXiv:2307.11019}, 2023.

\bibitem{10.1561/1500000019}
Stephen Robertson and Hugo Zaragoza.
\newblock The probabilistic relevance framework: Bm25 and beyond.
\newblock {\em Found. Trends Inf. Retr.}, 3(4):333–389, apr 2009.

\bibitem{sachan2021end}
Devendra~Singh Sachan, Mostofa Patwary, Mohammad Shoeybi, Neel Kant, Wei Ping, William~L Hamilton, and Bryan Catanzaro.
\newblock End-to-end training of neural retrievers for open-domain question answering.
\newblock {\em arXiv preprint arXiv:2101.00408}, 2021.

\bibitem{shen2024exploring}
Tingjia Shen, Hao Wang, Jiaqing Zhang, Sirui Zhao, Liangyue Li, Zulong Chen, Defu Lian, and Enhong Chen.
\newblock Exploring user retrieval integration towards large language models for cross-domain sequential recommendation.
\newblock {\em arXiv preprint arXiv:2406.03085}, 2024.

\bibitem{shi2023replug}
Weijia Shi, Sewon Min, Michihiro Yasunaga, Minjoon Seo, Rich James, Mike Lewis, Luke Zettlemoyer, and Wen-tau Yih.
\newblock Replug: Retrieval-augmented black-box language models.
\newblock {\em arXiv preprint arXiv:2301.12652}, 2023.

\bibitem{DBLP:conf/emnlp/0001PCKW21}
Kurt Shuster, Spencer Poff, Moya Chen, Douwe Kiela, and Jason Weston.
\newblock Retrieval augmentation reduces hallucination in conversation.
\newblock In Marie{-}Francine Moens, Xuanjing Huang, Lucia Specia, and Scott~Wen{-}tau Yih, editors, {\em Findings of the Association for Computational Linguistics: {EMNLP} 2021, Virtual Event / Punta Cana, Dominican Republic, 16-20 November, 2021}, pages 3784--3803. Association for Computational Linguistics, 2021.

\bibitem{Sun_Wang_Li_Feng_Tian_Wu_Wang_2020}
Yu~Sun, Shuohuan Wang, Yukun Li, Shikun Feng, Hao Tian, Hua Wu, and Haifeng Wang.
\newblock Ernie 2.0: A continual pre-training framework for language understanding.
\newblock {\em Proceedings of the AAAI Conference on Artificial Intelligence}, 34(05):8968--8975, Apr. 2020.

\bibitem{topsakal2023creating}
Oguzhan Topsakal and Tahir~Cetin Akinci.
\newblock Creating large language model applications utilizing langchain: A primer on developing llm apps fast.
\newblock In {\em Proceedings of the International Conference on Applied Engineering and Natural Sciences, Konya, Turkey}, pages 10--12, 2023.

\bibitem{touvron2023llama}
Hugo Touvron, Thibaut Lavril, Gautier Izacard, Xavier Martinet, Marie-Anne Lachaux, Timoth{\'e}e Lacroix, Baptiste Rozi{\`e}re, Naman Goyal, Eric Hambro, Faisal Azhar, et~al.
\newblock Llama: Open and efficient foundation language models.
\newblock {\em arXiv preprint arXiv:2302.13971}, 2023.

\bibitem{wang2023rfid}
Cunxiang Wang, Haofei Yu, and Yue Zhang.
\newblock Rfid: Towards rational fusion-in-decoder for open-domain question answering.
\newblock {\em arXiv preprint arXiv:2305.17041}, 2023.

\bibitem{wang2021hypersorec}
Hao Wang, Defu Lian, Hanghang Tong, Qi~Liu, Zhenya Huang, and Enhong Chen.
\newblock Hypersorec: Exploiting hyperbolic user and item representations with multiple aspects for social-aware recommendation.
\newblock {\em ACM Transactions on Information Systems (TOIS)}, 40(2):1--28, 2021.

\bibitem{wei2021finetuned}
Jason Wei, Maarten Bosma, Vincent~Y Zhao, Kelvin Guu, Adams~Wei Yu, Brian Lester, Nan Du, Andrew~M Dai, and Quoc~V Le.
\newblock Finetuned language models are zero-shot learners.
\newblock {\em arXiv preprint arXiv:2109.01652}, 2021.

\bibitem{DBLP:conf/iclr/WeiBZGYLDDL22}
Jason Wei, Maarten Bosma, Vincent~Y. Zhao, Kelvin Guu, Adams~Wei Yu, Brian Lester, Nan Du, Andrew~M. Dai, and Quoc~V. Le.
\newblock Finetuned language models are zero-shot learners.
\newblock In {\em The Tenth International Conference on Learning Representations, {ICLR} 2022, Virtual Event, April 25-29, 2022}. OpenReview.net, 2022.

\bibitem{wu2024survey}
Likang Wu, Zhi Zheng, Zhaopeng Qiu, Hao Wang, Hongchao Gu, Tingjia Shen, Chuan Qin, Chen Zhu, Hengshu Zhu, Qi~Liu, et~al.
\newblock A survey on large language models for recommendation.
\newblock {\em World Wide Web}, 27(5):60, 2024.

\bibitem{xiao-etal-2021-ernie}
Dongling Xiao, Yu-Kun Li, Han Zhang, Yu~Sun, Hao Tian, Hua Wu, and Haifeng Wang.
\newblock {ERNIE}-gram: Pre-training with explicitly n-gram masked language modeling for natural language understanding.
\newblock In {\em Proceedings of the 2021 Conference of the North American Chapter of the Association for Computational Linguistics: Human Language Technologies}, pages 1702--1715, Online, June 2021. Association for Computational Linguistics.

\bibitem{DBLP:conf/naacl/YangXLLTXLL19}
Wei Yang, Yuqing Xie, Aileen Lin, Xingyu Li, Luchen Tan, Kun Xiong, Ming Li, and Jimmy Lin.
\newblock End-to-end open-domain question answering with bertserini.
\newblock In Waleed Ammar, Annie Louis, and Nasrin Mostafazadeh, editors, {\em Proceedings of the 2019 Conference of the North American Chapter of the Association for Computational Linguistics: Human Language Technologies, {NAACL-HLT} 2019, Minneapolis, MN, USA, June 2-7, 2019, Demonstrations}, pages 72--77. Association for Computational Linguistics, 2019.

\bibitem{yang2018hotpotqa}
Zhilin Yang, Peng Qi, Saizheng Zhang, Yoshua Bengio, William~W Cohen, Ruslan Salakhutdinov, and Christopher~D Manning.
\newblock Hotpotqa: A dataset for diverse, explainable multi-hop question answering.
\newblock {\em arXiv preprint arXiv:1809.09600}, 2018.

\bibitem{yao2024seakrselfawareknowledgeretrieval}
Zijun Yao, Weijian Qi, Liangming Pan, Shulin Cao, Linmei Hu, Weichuan Liu, Lei Hou, and Juanzi Li.
\newblock Seakr: Self-aware knowledge retrieval for adaptive retrieval augmented generation, 2024.

\bibitem{yasunaga-etal-2022-linkbert}
Michihiro Yasunaga, Jure Leskovec, and Percy Liang.
\newblock {L}ink{BERT}: Pretraining language models with document links.
\newblock In {\em Proceedings of the 60th Annual Meeting of the Association for Computational Linguistics (Volume 1: Long Papers)}, pages 8003--8016, Dublin, Ireland, May 2022. Association for Computational Linguistics.

\bibitem{yin2024dataset}
Mingjia Yin, Hao Wang, Wei Guo, Yong Liu, Suojuan Zhang, Sirui Zhao, Defu Lian, and Enhong Chen.
\newblock Dataset regeneration for sequential recommendation.
\newblock In {\em Proceedings of the 30th ACM SIGKDD Conference on Knowledge Discovery and Data Mining}, pages 3954--3965, 2024.

\bibitem{yu2023augmentation}
Zichun Yu, Chenyan Xiong, Shi Yu, and Zhiyuan Liu.
\newblock Augmentation-adapted retriever improves generalization of language models as generic plug-in.
\newblock {\em arXiv preprint arXiv:2305.17331}, 2023.

\bibitem{DBLP:conf/emnlp/ZhongLC22}
Zexuan Zhong, Tao Lei, and Danqi Chen.
\newblock Training language models with memory augmentation.
\newblock In Yoav Goldberg, Zornitsa Kozareva, and Yue Zhang, editors, {\em Proceedings of the 2022 Conference on Empirical Methods in Natural Language Processing, {EMNLP} 2022, Abu Dhabi, United Arab Emirates, December 7-11, 2022}, pages 5657--5673. Association for Computational Linguistics, 2022.

\bibitem{zhu2021retrieving}
Fengbin Zhu, Wenqiang Lei, Chao Wang, Jianming Zheng, Soujanya Poria, and Tat-Seng Chua.
\newblock Retrieving and reading: A comprehensive survey on open-domain question answering.
\newblock {\em arXiv preprint arXiv:2101.00774}, 2021.

\end{thebibliography}
